\documentclass[letterpaper, 10 pt, conference]{src/ieeeconf}  

\IEEEoverridecommandlockouts                              
\usepackage{graphicx}
\usepackage{amsmath,amssymb}
\usepackage{cite}
\usepackage{float}
\usepackage[caption=false,font=footnotesize]{subfig}

\graphicspath{{img/},{fig/}}

\title{\LARGE \bf
A practical optimal control approach for two-speed actuators
}

\author{Alexandre Girard and H. Harry Asada
\thanks{This work was supported in part by The Boeing Company, the Fonds Qu\'{e}b\'{e}cois de la recherche sur la nature et les technologies (FQRNT) and the Natural Sciences and Engineering Research Council of Canada (NSERC). }%
\thanks{A. Girard and H. H. Asada are with Department of Mechanical Engineering, Massachusetts Institute of Technology, Cambridge, MA, USA, email : \{agirard,asada\}@mit.edu }%
}

\begin{document}

\maketitle
\thispagestyle{empty}
\pagestyle{empty}

\begin{abstract}
This paper addresses the closed-loop control of an actuator with both a continuous input variable (motor torque) and a discrete input variable (mode selection). In many applications, robots have to bear large loads while moving slowly and also have to move quickly through the air with almost no load, leading to conflicting requirements for their actuators. An actuator with multiple gear ratios, like in a powertrain, can address this issue by allowing an effective use of power over a wide range of output speed. However, having discrete modes of operation adds complexity to the high-level control and planning. Here a controller for two-speed actuators that automatically select both the best gear ratio and the motor torque is developed. The approach is to: first derive a low-dimensional model, then use dynamic programming to find the best actions for all possible situations, and last use regression analysis to extract simplified global feedback laws. This approach produces simple practical nearly-optimal feedback laws. A controller that globally minimizes a quadratic cost function is derived for a two-speed actuator prototype, global stability is proven and performance is demonstrated experimentally.
\end{abstract}


\section{INTRODUCTION}

Robotic systems often operate in distinctively different torque-speed conditions. For instance, a legged robot has to move its leg forward quickly through the air and, once touching the ground, it has to bear a large load \cite{hirose_study_1984}. Those challenging extremum requirements often lead to the use of oversized and inefficient actuators, which inhibit the performance of robots that carry their actuators and energy source \cite{girard_two-speed_2015} \cite{Wolfgang_novel_2015}, like wearable and mobile robots. Actuators using multiple gear ratios, like in a powertrain, can address this issue by allowing a more effective use of power over a wide range of speeds. However, gear-shifting is a very non-linear process and its use in a robotic context raise many issues. One challenge is the high-level control of a system with discrete modes of operation. The issue comes from the fact that the system has input variables of different natures: continuous variables (motor torques) and discrete variables (mode selections), hence most control engineering tools are not suited to tackle this problem. 

\begin{figure}[htpb]
				\vspace{-10pt}
        \centering
				\subfloat[Two-speed actuator]{
				\includegraphics[width=0.20\textwidth]{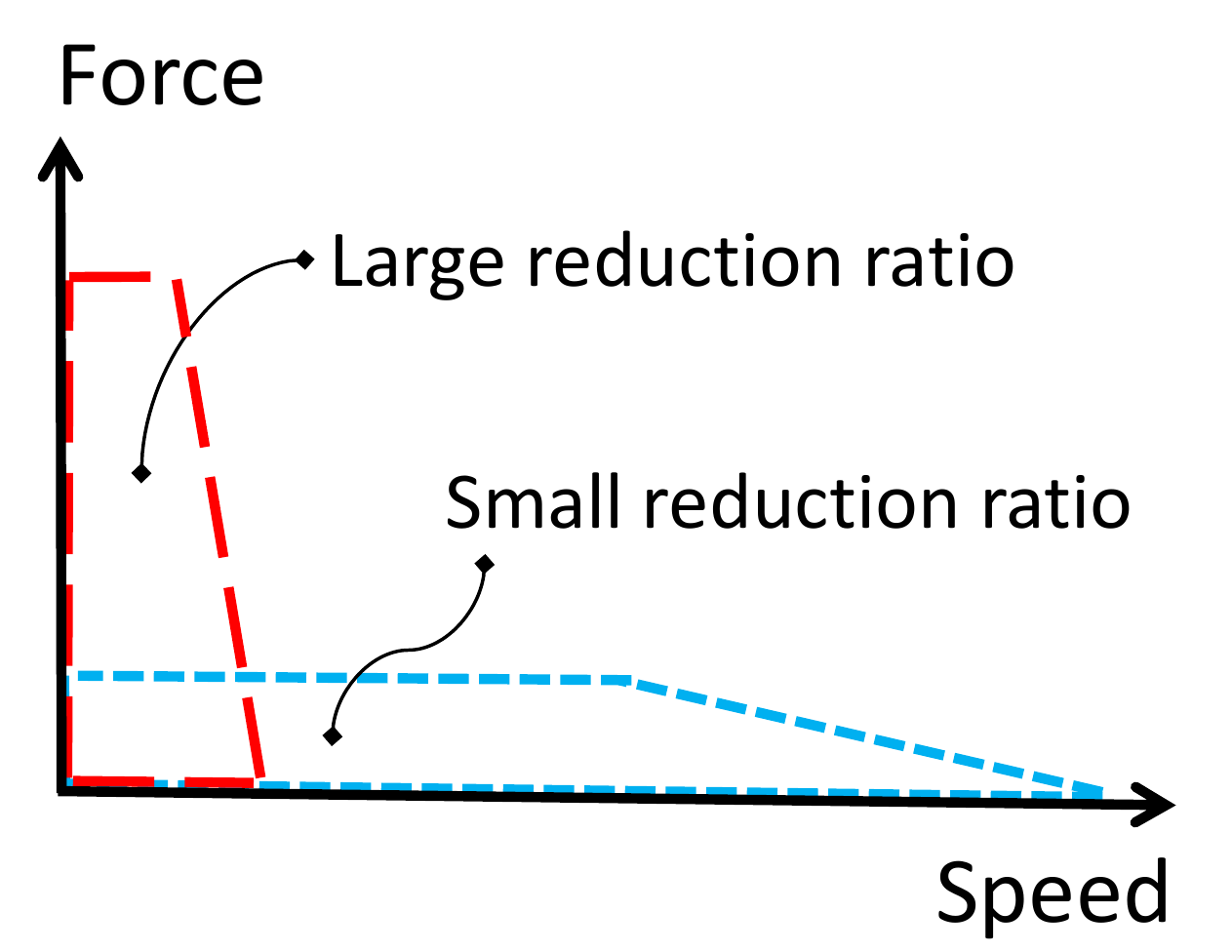}
				\label{fig:fs}}
        \subfloat[2 DOF manipulator ]{
				\includegraphics[width=0.25\textwidth]{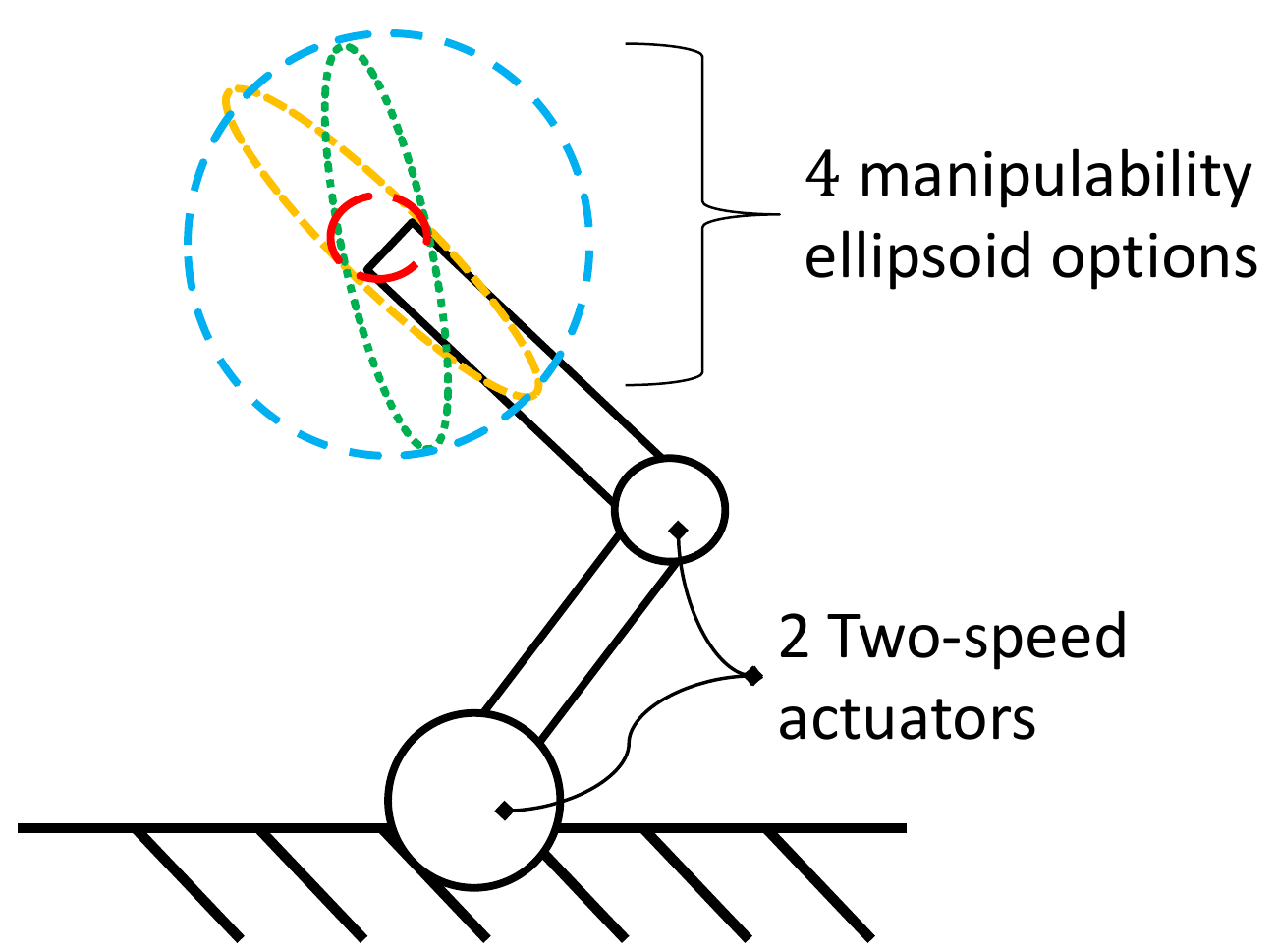}
				\label{fig:2DOF}}
        \caption{Force-speed characteristics of different operating modes}
\end{figure}

For simple scenarios like steady state operations, the mode selection can be based on simple principles. For instance, for a two-speed electric motor at a steady speed, the best operating mode for power output could be selected based on torque-speed curves (see Fig. \ref{fig:fs}). Alternatively, for maximizing accelerations, the reduction ratio could be selected based on the best motor-load inertia match. However, finding the best mode selection policy for a general task where the whole trajectory is important cannot be done using such point-by-point optimums. For instance, to globally minimize the energy use of a car, it can be advantageous to coast with the transmission in the neutral mode and use the engine only intermittently. To find such solutions the global behavior and whole trajectories must be considered. Furthermore, the problem gets harder for robotic systems because of the multi-DOF aspects. For instance, a manipulator using two two-speed actuators would have $2^2$ operating modes, with their force properties illustrated by 4 possible manipulability ellipsoids at Fig. \ref{fig:2DOF}. Hence, except for some basic scenario it is not trivial to find good feedback policies for the mode selection in a robotics context and an approach generalizable to complex situations is thus needed.

\begin{figure}[b]
	\centering
	\vspace{-10pt}
		\includegraphics[width=0.40\textwidth]{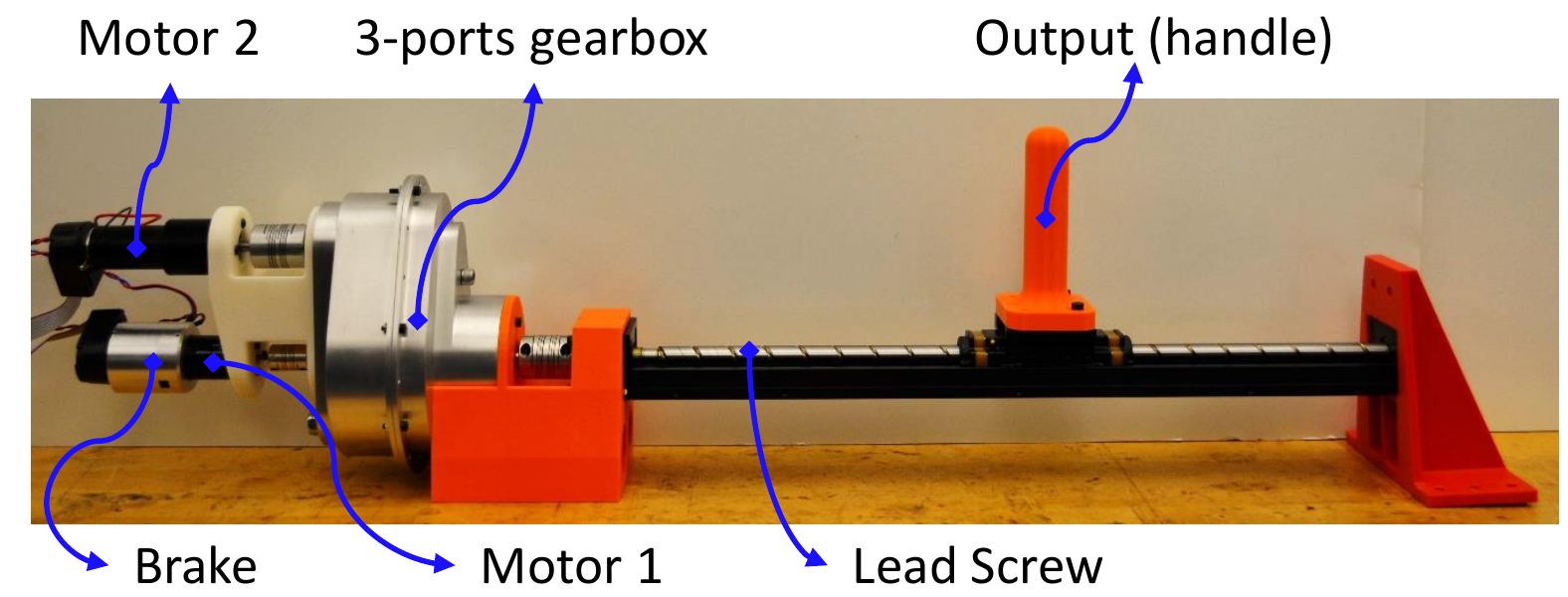}
	\caption{Dual-speed dual-motor linear actuator prototype}
	\label{fig:proto_linear}
\end{figure}

\begin{figure}[ht]
	\centering
		\includegraphics[width=0.45\textwidth]{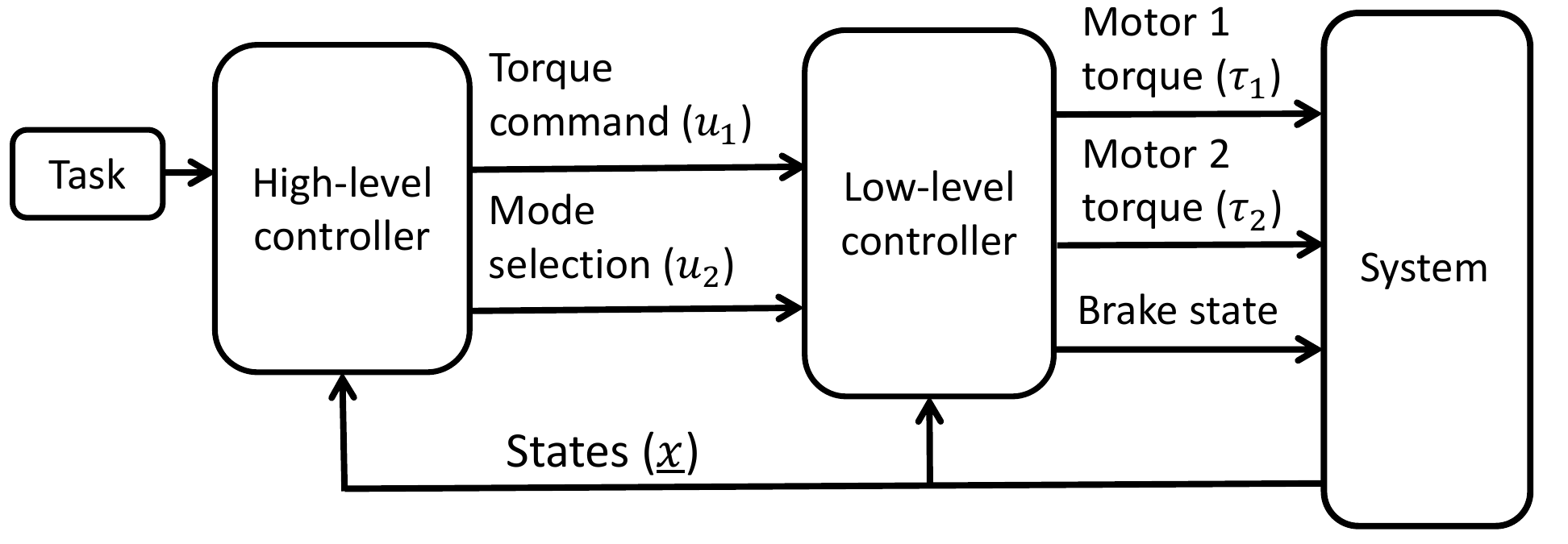}
	\caption{Controller architecture}
	\label{fig:blocks}
\end{figure}

This paper address the design of global feedback laws, for both continuous and discrete inputs, that lead to a desired global behavior, which will be formulated as reaching a target while minimizing a cost function. A practical computational approach that put simplicity before absolute optimality is proposed. The technique could be used on any low-dimensional system for a wide range of control objectives. Here, this paper focus on demonstrating the approach for the task of driving a 11.4 kg (25 lbs) mass for point-to-point motions using the two-speed actuator shown at Fig. \ref{fig:proto_linear}. 
This actuator use a special dual-motor architecture to enable fast and seamless gear shifting, which is critical in robotic applications. Motor 1 and motor 2  have reduction ratios of 4:1 and 72:1 respectively and are coupled to a ball screw (20 mm lead) through a 3-ports gearbox. The motor in use can be selected by closing or opening the brake. In previous work, low-level controllers were developed to control each mode of operation independently and transitions from one mode to another \cite{girard_two-speed_2015}. In this paper, a high-level controller that selects the operating mode and the torque command is designed (see Fig. \ref{fig:blocks}). The organization of the paper is as follow: In section \ref{sec:RelatedWorks}, related control techniques are briefly reviewed. In section \ref{sec:cls}, global feedback laws for the two-speed actuator are derived using dynamic programming. In section \ref{ev}, experimentation results using the prototype are presented and discussed. 

\subsection{Related works}
\label{sec:RelatedWorks}

Discrete mode of operations are present in many system, such as powertrains with multiple gear ratios, pneumatic or hydraulic systems equipped with on/off valves and special actuators \cite{leach_linear_2012}\cite{Wolfgang_novel_2015}\cite{lee_finger_2013}. Considering the global behavior of such systems, including the mode selection, lead to a hybrid dynamical model. Most optimal control techniques are based on either variational approaches or some form of gradient descent to find a trajectory that minimize a cost function \cite{betts_practical_2010}. Hence those techniques cannot be used directly to optimize discrete variables. An interesting approach to get around this problem is to use the switching instants as optimization parameters instead \cite{xu_optimal_2004}\cite{majdoub_hybrid_2010}. However, to use this approach a sequence of operating modes must be predefined first. Mixed-integer programming has also been used to generate optimal open-loop trajectories of dynamical system with both continuous and discrete input variables \cite{richards_spacecraft_2002} \cite{gerdts_solving_2005}. However, open-loop trajectories can be unstable and new trajectories must be computed for each initial condition. Hence, from a practical point of view, techniques generating feedback policies are preferable. Most of the analytical results regarding feedback control of hybrid systems are for specific cases, for instance the optimal feedback law of linear hybrid systems with linear constraints for a quadratic cost function have been shown to have a particular form \cite{borrelli_dynamic_2005}. One computational technique that generate feedback laws and that can be used for non-linear systems with any kind of constraints is dynamic programming \cite{donald_e._kirk_optimal_2004}. Two disadvantages of the techniques are however that it only works for low-dimensional systems (so called curse of dimensionality) and also that the resulting feedback laws are in the form of a look-up table.


\section{Control laws synthesis}
\label{sec:cls}

\subsection{Methodology overview}

The proposed approach is to discretize the continuous control problem into a graph search problem where the discrete input actions can be considered naturally, and then use dynamic programming to find global optimal feedback policies. First, a simplified low-dimensional model of the controlled system is derived to keep the problem tractable. Second, the optimal control problem is solved using dynamic programming. Third, the resulting optimal policies are approximated with simple feedback laws, to conduct a stability analysis and also for the ease of implementation.

\subsection{Modeling}
\label{sec:Modeling}

\begin{figure}[H]
	\centering
		\includegraphics[width=0.42\textwidth]{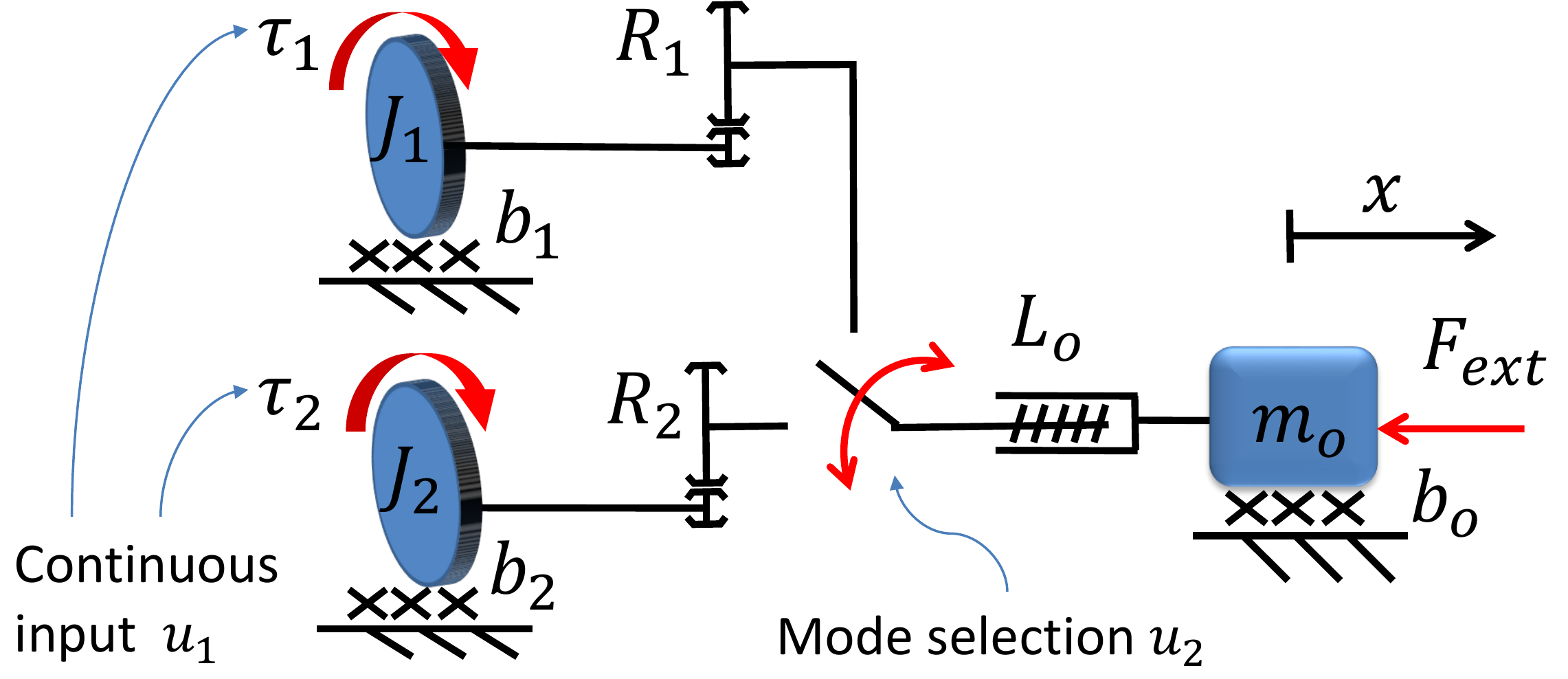}
	\caption{Lumped parameter simplified model}
	\label{fig:model}
\end{figure}

Using the previously developed low-level controller, when the selected mode is changed the system automatically transit from one mode to another quickly and seamlessly \cite{girard_two-speed_2015}. Hence, for the purpose of designing the high-level controller, a simple model that assumes transitions are instantaneous is used, see Fig. \ref{fig:model}. While this model does not capture the transition behavior, it has the advantage of capturing most of the global behavior with only two state variables. However, this modeling simplification can be unsuited for loads exhibiting fast dynamics
%
Two input variables are used; $u_2 \in \{ 1,2 \}$ a discrete variable representing the mode selection and $u_1 \in \Re^1$ representing the continuous torque command sent to the motor currently in use. Hence, the model is formulated with a hybrid form:
\begin{align}
	\ddot{x} =& f(\dot{x},u_1,u_2) =   \textstyle \frac{1}{m_{r,i}} \left[ - b_{r,i} \dot{x}  +  \frac{R_i}{L_o}  u_1 \right] \displaystyle , \; i=u_2  
	\label{eq:ddx}
\end{align}
where $m_{r,i}$ and $b_{r,i}$ are the reflected mass and damping of each mode of operation:
\begin{align}
\textstyle m_{r,i} = m_o + \left( \frac{R_i}{L_o} \right)^2 J_i  \quad \quad  b_{r,i} = b_o + \left( \frac{R_i}{L_o} \right  )^2 b_i 
\label{eq:reflected}
\end{align}
In state-space form:
\begin{align}
\textstyle
	\underline{\dot{x}} = F(\underline{x}, \underline{u}) =
	\left[ \begin{array}{c} 
	\dot{x} \\ 
	\text{eq.(1)} 
	\end{array}\right]	
	\quad \underline{x} = \left[ \begin{array}{c} 	x \\ 	\dot{x}  	\end{array} \right]  \;
	\underline{u} = \left[ \begin{array}{c} 	u_1 \\ 	u_2  	\end{array} \right]
	\label{eq:ss}
\end{align}

\subsection{Problem formulation}
\label{sec:ProblemFormulation}

The control problem of obtaining the global desired behavior is formulated as minimizing a scalar cost $J$ that is a function of the state trajectory $\underline{x}(:)$ and the inputs trajectory $\underline{u}(:)$, while constraining both states and inputs to be in their respective domains:
\begin{align}
	\operatornamewithlimits{min}\limits_{\underline{u}(:)} \; & J(\underline{x}(:),\underline{u}(:)) = \int g(\underline{x}(t),\underline{u}(t)) dt  \\
	s.t. \quad & \underline{\dot{x}} = F(\underline{x}, \underline{u}) \\
	& u_1 \in [ -0.02 , 0.02 ] \quad (Nm) \\
	& u_2 \in \{ 1 , 2 \} \\
	& x_1 \in [ -150 , 150 ] \quad (mm) \\
	& x_2 \in [ -500 , 500 ] \quad (mm/sec)
	\label{eq:min}
\end{align}

An additional constraint is imposed, when the output speed is greater than 20 mm/sec only the operating mode $u_1=1$ can be used (to limit motor speeds to 8000 RPM). Then three different additive cost functions are investigated; a quadratic cost function $g_q$ where error is weighted against control effort, a function corresponding $g_t$ to the minimal time problem and a function $g_e$ corresponding to thermal losses in the motors, leading to a simplified minimum wasted energy problem (neglecting mechanical losses).
\begin{align}
	g_q(\underline{x},\underline{u}) &= w_1 \; x^2 + w_2 \; \dot{x}^2 + w_3 \; u_1^2
	\label{eq:g_quad} \\
	g_t(\underline{x},\underline{u}) &= 1
	\label{eq:g_time} \\
	g_e(\underline{x},\underline{u}) &= u_1^2 \quad  \propto \quad  \dot{Q}_{motor} = r I^2
	\label{eq:g_e}
\end{align}

\subsection{Dynamic programming parameters}
\label{sec:DynamicPrograming}

A dynamic programming algorithm, also known as value iteration, is used to find both the optimal cost-to-go function and optimal policies for an infinite horizon. The optimal control problem is numerically solved for the different three cost functions. The discretization parameters are as follow: the time step is 0.02 sec, the state space is discretized into an even 501 x 501 grid and the continuous torque is discretized into 51 discrete control options, for a total of 102 possible control actions including the mode selection. Artificial out-of-bound and on-target termination states are included to guarantee convergence \cite{DP}.


\subsection{Numerical results}
\label{sec:NumericalResults}
\begin{figure*}[htpb]
        \centering
				\subfloat[Minimum time]{
        \includegraphics[width=0.32\textwidth]{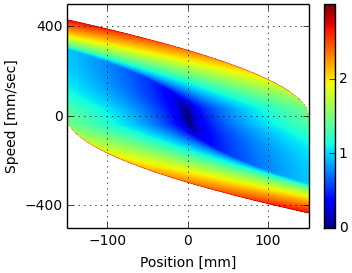}
				\label{fig:J_time}}
        \subfloat[Quadratic cost]{
				\includegraphics[width=0.32\textwidth]{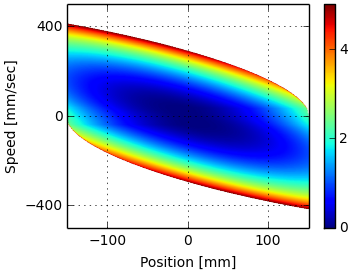}
				\label{fig:J_LQR}}
				\subfloat[Minimum energy]{
				\includegraphics[width=0.32\textwidth]{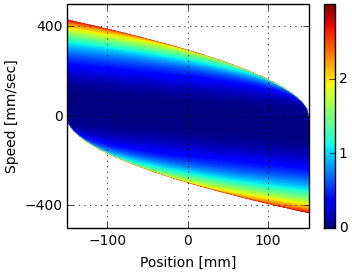}
				\label{fig:J_energy}}
        \caption{Optimal cost-to-go $J^*$}\label{fig:J}
\end{figure*}

\begin{figure*}[htpb]
        \centering
				\subfloat[Minimum time]{
        \includegraphics[width=0.32\textwidth]{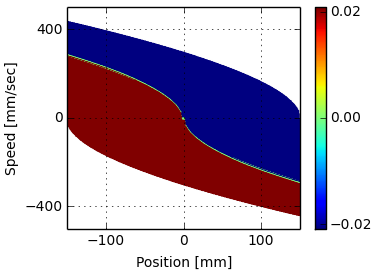}
				\label{fig:u0_time}}
        \subfloat[Quadratic cost]{
				\includegraphics[width=0.32\textwidth]{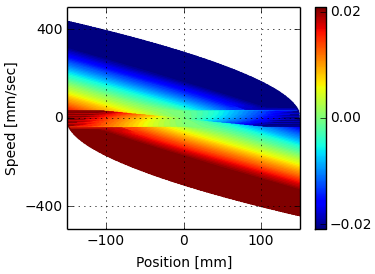}
				\label{fig:u0_LQR}}
				\subfloat[Minimum energy]{
				\includegraphics[width=0.32\textwidth]{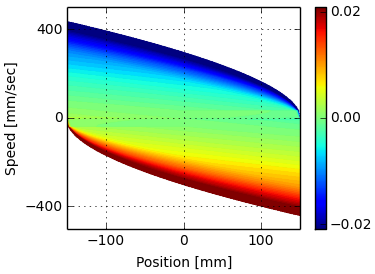}
				\label{fig:u0_energy}}
        \caption{Optimal policy for the continuous torque command $u_1$ [Nm]}\label{fig:u0}
\end{figure*}

\begin{figure*}[htpb]
        \centering
				\subfloat[Minimum time]{
        \includegraphics[width=0.32\textwidth]{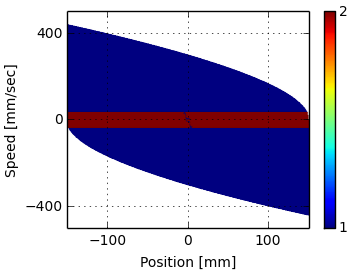}
				\label{fig:u1_time}}
        \subfloat[Quadratic cost]{
				\includegraphics[width=0.32\textwidth]{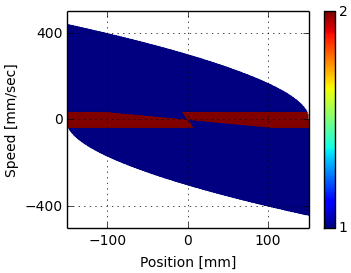}
				\label{fig:u1_LQR}}
				\subfloat[Minimum energy]{
				\includegraphics[width=0.32\textwidth]{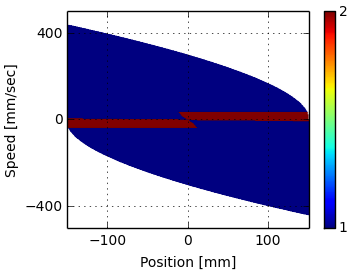}
				\label{fig:u1_energy}}
        \caption{Optimal policy for the mode selection $u_2$}\label{fig:u1}
\end{figure*}

\begin{figure*}[htpb]
        \centering
				\subfloat[Minimum time]{
        \includegraphics[width=0.32\textwidth]{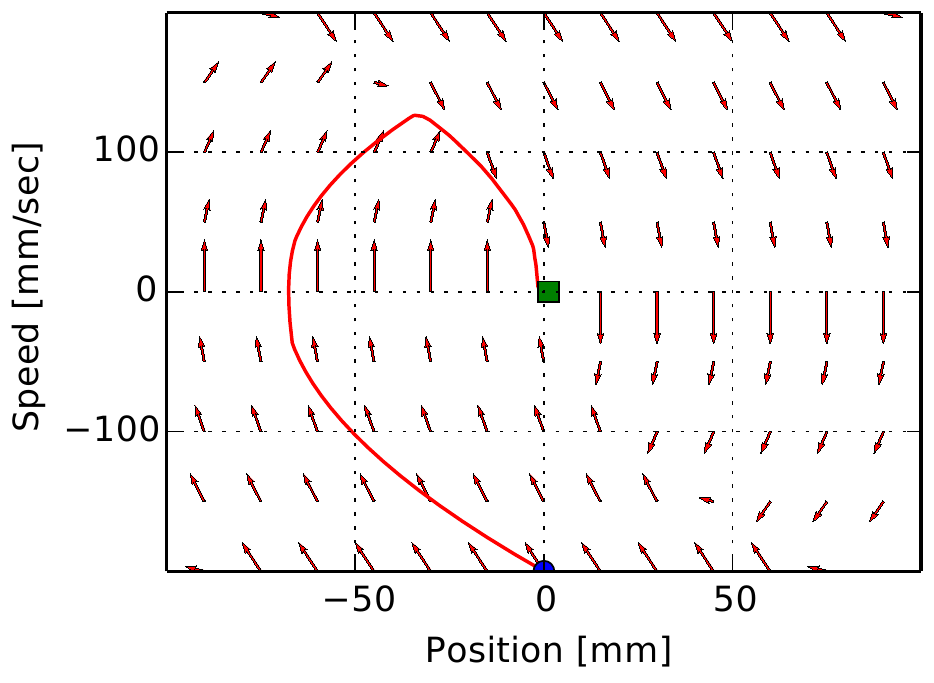}
				\label{fig:phase_plane_time}}
        \subfloat[Quadratic cost]{
				\includegraphics[width=0.32\textwidth]{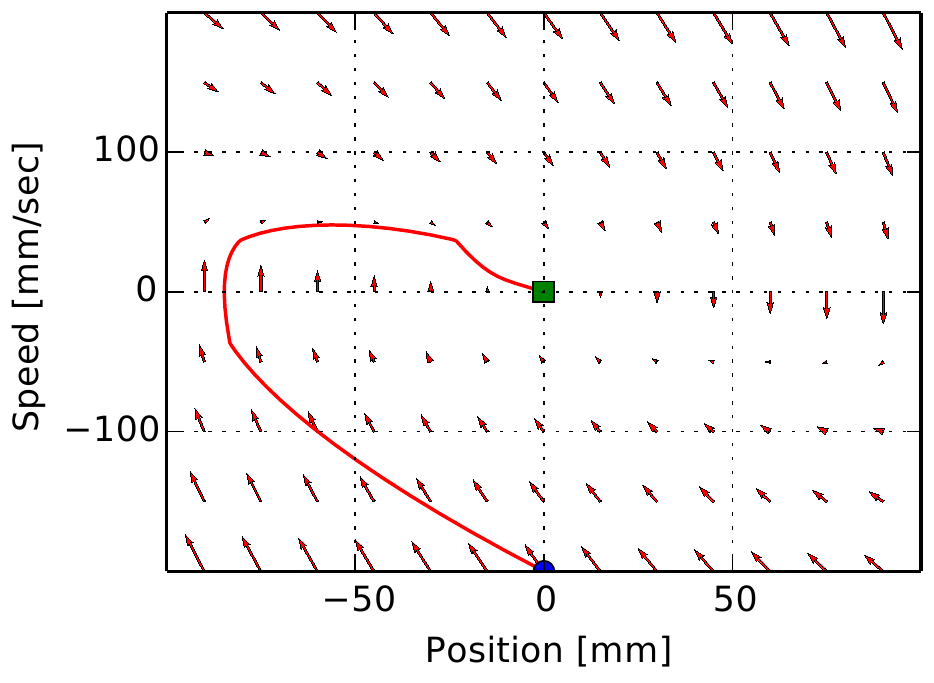}
				\label{fig:phase_plane_LQR}}
				\subfloat[Minimum energy]{
				\includegraphics[width=0.32\textwidth]{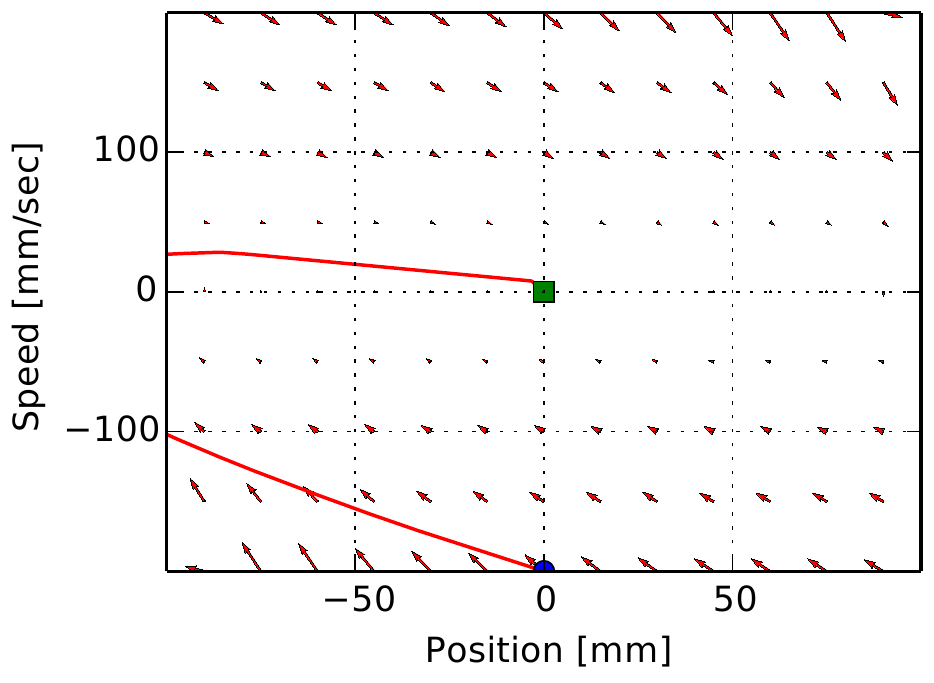}
				\label{fig:phase_plane_energy}}
        \caption{Closed loop behavior with the optimal policy illustrated in the phase plane}\label{fig:phase_plane}
\end{figure*}

Fig. \ref{fig:J}-\ref{fig:u1} illustrate the numerical results of the dynamic programming algorithm for the three different cost functions. The absence of color indicates states with no solution (a constraint will be violated for any possible control actions). Fig. \ref{fig:phase_plane} shows the closed loop behavior of the system in the phase plane when the optimal policy is applied.


\paragraph{Minimum time}
\label{sec:MinimumTime}
For the minimum time problem, the optimal policy is a bang-bang law for $u_1$ and always using highly-geared mode when possible. Note that the bang-bang switching curve accounts for the fact the large gear ratio will be used during the final part of the trajectory.

\paragraph{Quadratic cost}
\label{sec:QuadraticCost}
For the quadratic cost, the mode selection optimal policy is almost as simple as the minimum time problem except for small features in quadrant II and IV. The more interesting result comes from the continuous torque control law, the gains when using the large reduction ratio are larger than those when using the small reduction ratio. This results in the controller taking action mainly at low speed when its actions have the biggest impacts on the system, and lead to a highly non-linear closed-loop behavior. This is the opposite of what would have been obtained using feedback linearization: large gains with the small gear and small gains with the large gear to linearize the global behavior. Hence implementing feedback linearization would have led to a poor performance considering this quadratic cost metric. 

\paragraph{Minimum energy}
\label{sec:MinimumEnergy}
For the minimum energy controller, interestingly the mode selection policy is not trivial even for this simple linear model.  This shows that it does not take much complexity to have non-trivial optimal policies for hybrid systems. Here, the large reduction ratio is used almost only for braking, and in quadrant II and IV the small reduction ratio is used even at low speed to coast with low viscous resistance. Also globally the gains are much lower than the other controllers except for zones where it is necessary to use energy to stay in the domain.

\subsection{Control laws approximation}
\label{sec:aprx}

In this section, simplified algebraic control laws are extracted from the numerical map of optimal control actions for the quadratic cost controller. The dynamic programming algorithm outputs a list of optimal actions that is assumed are samples of a  map of optimal actions $\underline{u}^* = \pi^* ( \underline{x} )$.
The goal is then to find a simplified map $\hat{\pi}( \underline{x} )$ based on the samples. Here a two-steps approach is used; first the state-space is segmented into zones based on the optimal discrete action $u_2$, then regression is used to compute the $u_1$ map independently in each zone. 

\setcounter{paragraph}{0}
\paragraph{Segmentation}
For the resulting optimal policy found at Fig. \ref{fig:u0_LQR}, three main zones can be identified. The boundaries are approximated by two lines at $\dot{x}= \pm \; 20 $, and the samples are separated into two groups:
\begin{align}
\left \{ \underline{u}^i , \underline{x}^i \right \} \rightarrow \text{Group 1} \quad \text{if} \; | \dot{x} | \geq 20 \\
\left \{ \underline{u}^i , \underline{x}^i \right \} \rightarrow \text{Group 2} \quad \text{if} \; | \dot{x} | < 20
\end{align}

For more complex segmentation, machine learning techniques could be used to perform the classification.

\paragraph{Continuous maps}

\begin{figure}[htpb]
				\vspace{-10pt}
        \centering
				\subfloat[Map for $u_1$ (Nm)]{
        \includegraphics[width=0.22\textwidth]{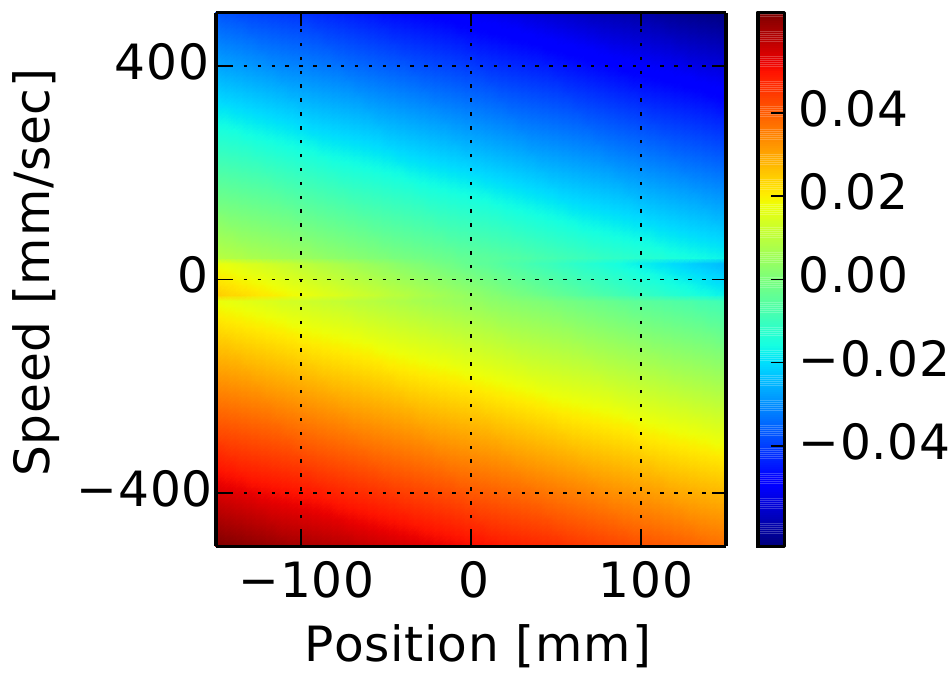}
				\label{fig:u0_LQR_approx}}
        \subfloat[Phase plane behavior]{
				\includegraphics[width=0.22\textwidth]{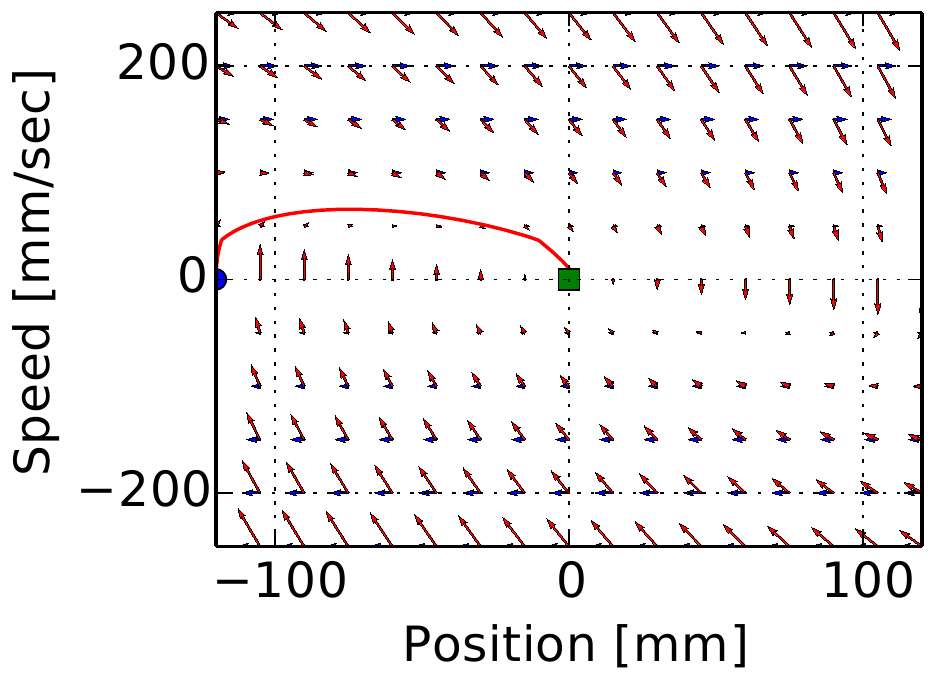}
				\label{fig:phase_plot_lqr}}
        \caption{Approximated control laws and resulting behavior}\label{fig:approx}
\end{figure}

In each zones, the optimal continuous input $u_1^*$ is very close to a planar surface with exceptions when the input saturates or near the boundaries, see Fig. \ref{fig:u0_LQR}. Here, it is proposed to forgo absolute optimality and instead use the simplest control law that can globally approximate $u_1^*$ map. Hence, simple linear plane are fitted on the samples group by group. The regression is defined as follow, the dependent variable to approximate is $u_1$, the independent variables are the state coordinates $\underline{x}^T = [ x  \; \dot{x} ]$ and the parameter vector is $\underline{\alpha}^T = [ k_p  \; k_d ]$. Hence the continuous control variable $\hat{u}_1$ is approximated with the following linear map:
\begin{align}
\hat{u}_1 = \underline{\alpha}^T \underline{x} = k_p  x + k_d \dot{x}
\end{align}
which is essentially a simple proportional-derivative control law. Then the parameters are estimated using a least-square criterion, $\underline{\alpha}_1$ using the samples in group 1 and $\underline{\alpha}_2$ samples in group 2. Then combining the segmentation and both linear maps, the global control policy is approximated with the following law:
\begin{align}
\underline{ \hat{u}} &= \hat{\pi} ( \underline{x} )
 = \left \{ 
	\begin{array}{c}
	\left[
	\begin{array}{c}
		 \underline{\alpha}_1^T \underline{x} \\
		 1 
	\end{array} 
	\right] 		
	\text{if} \; | \dot{x} | \geq 20 \\ \\
		 \left[
		\begin{array}{c}
		 \underline{\alpha}_2^T \underline{x} \\ 
		 2
	\end{array}
		 \right]
		\text{if} \; | \dot{x} | < 20
	\end{array}
	\right.
	\label{eq:uhat}
\end{align}
Fig. \ref{fig:u0_LQR_approx} shows the resulting continuous control policy $\hat{u}_1$ for the full state-space, and is an approximation of Fig. \ref{fig:u0_LQR} map.


The resulting behavior with the control law of eq.\eqref{eq:uhat} is illustrated in the phase plane at Fig. \ref{fig:phase_plot_lqr}, where the arrows illustrate the state derivative throughout the state space. The blue arrows illustrate the natural dynamic of the system and the red arrows illustrate the closed behavior with the control policy of eq.\eqref{eq:uhat}. At high speed with the small reduction ratio, the controller only takes small actions illustrated by the fact that red arrows only have small deviation compared to blue arrows. However at low speed, the large reduction ratio is used to drastically change the natural behavior of the system, illustrated by the large red arrows. 


\subsection{Stability analysis}
\label{st}

\begin{figure}[htpb]
        \centering
				\subfloat[Phase plane trajectory]{
        \includegraphics[width=0.22\textwidth]{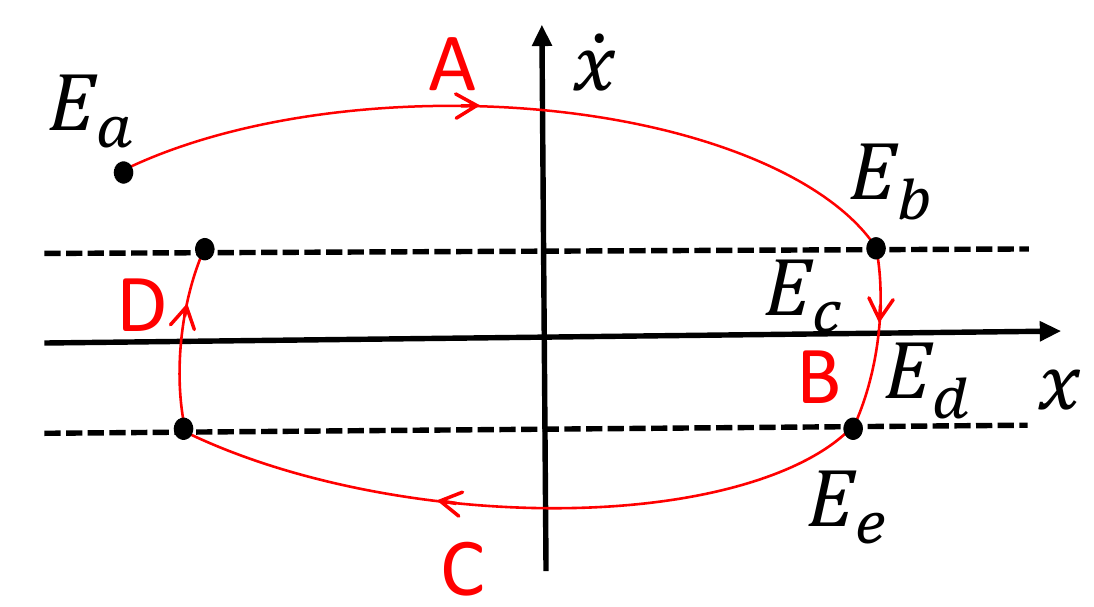}
				\label{fig:stability}}
        \subfloat[Time trajectory]{
				\includegraphics[width=0.22\textwidth]{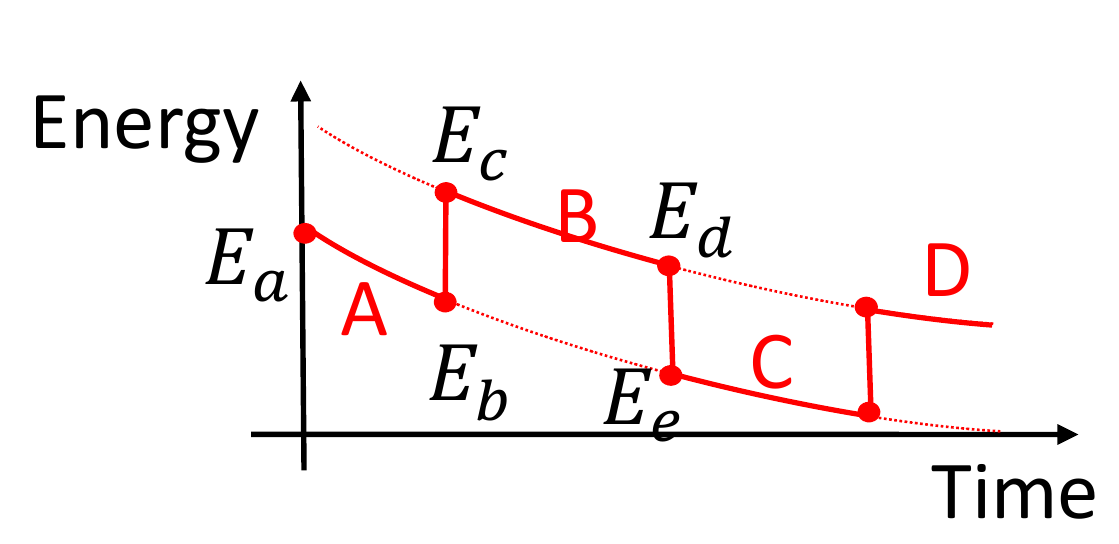}
				\label{fig:energy}}
        \caption{Energy along a trajectory}\label{fig:energytraj}
\end{figure}

Here in this section, the global stability of the controller generated for the quadratic cost is demonstrated. When the two-speed actuator is controlled with the feedback law of eq.\eqref{eq:uhat}, the closed loop behavior is given by:
\begin{align}
	\ddot{x} &= 
	\left \{ 
	\begin{array}{c}
	\frac{1}{m_{r,1}} \left[ - ( b_{r,1} + \frac{R_1}{L_o} k_{d1} ) \dot{x} - \frac{R_1}{L_o} k_{p1} x \right]  \quad \text{if} \; | \dot{x} | \geq 20 \\
	\frac{1}{m_{r,2}} \left[ - ( b_{r,2} + \frac{R_2}{L_o} k_{d2} ) \dot{x} - \frac{R_2}{L_o} k_{p2} x \right]  \quad \text{if} \; | \dot{x} | <    20
	\end{array}
	\right.
	\label{eq:closed_loop}
\end{align}
It will be shown that the energy (including virtual energy in the controller) of the system is always decreasing. The energy function is discontinuous across the switching surfaces, equal to $E_1$ when $| \dot{x} | \geq 20$ and $E_2$ when $| \dot{x} | < 20$, with value and time derivative given by:
\begin{align}
E_i        &= \frac{1}{2} \bigg[ m_{r,i} \bigg] \dot{x}^2 + \frac{1}{2} \left[ \frac{R_i}{L_o} k_{p,i} \right] x^2  &i=\{1,2\} \\
\dot{E}_i  &= -\left[ b_{r,i} + \frac{R_i}{L_o} k_{d,i} \right] \dot{x}^2                                           &i=\{1,2\}
	\label{eq:energy}
\end{align}
If $k_{p,i}$ and $(b_{r,i} + \frac{R_i}{L_o} k_{d,i})$ are positive values then $E_i > 0$ and $\dot{E}_i < 0$ for any non-zero point in the state space.
%
Hence all trajectories staying in their respective zone have decreasing energy. 
Fig. \ref{fig:stability} show a general trajectory on the phase plane, and Fig. \ref{fig:energy} shows the corresponding time-trajectory of the energy. Taken independently the energy is decreasing on trajectory segment A ($E_b<E_a$) and B ($E_d<E_c$). However, when crossing a switching surface the energy can increase, for instance $E_c>E_b$. For a down shift the kinetic energy increase as the output speed stay the same but the reflected inertia is greater with the larger reduction ratio. Note that this is not an artifact of using a simplified actuator model: the motor needs to be accelerated to match the output speed during the gear shift process. 
%
%
To prove global stability, it must be shown that the energy cannot increase by going back and forth into different operating modes. Hence, that $E_e<E_b$ for any trajectory. First a mapping $E_2 = \textit{h}(E_1)$ of the energy change when crossing the switching surface is needed. On the crossing surface the energy is only a function of the position:
\begin{align}
E_i(x)       &= \frac{1}{2} \bigg[ m_{r,i} \bigg] s^2 + \frac{1}{2} \left[ \frac{R_i}{L_o} k_{p,i} \right] x^2  
\end{align}
where $s$ is the speed on the switching surface. The energy correspondence is:
\begin{align}
E_2 = \textit{h}(E_1) = \frac{1}{2} \bigg[ \scriptstyle m_{r,2} - \frac{R_2 k_{p,2} }{R_1 k_{p,1}} \bigg] \displaystyle s^2 + 
\bigg[ \scriptstyle\frac{R_2 k_{p,2} }{R_1 k_{p,1}} \bigg] \displaystyle E_1
\label{eq:map}
\end{align}
Then it is possible to prove that $E_e<E_b$ for any trajectory starting with the inequality $E_d<E_c$ and substituting with the mapping of eq.\eqref{eq:map}:
\begin{align}
E_d < E_c  \; \Rightarrow \;  \textit{h}(E_e) < \textit{h}(E_b)  \; \Rightarrow \; E_e <  E_b  
\end{align}

Hence, any trajectory going into the low-speed zone and coming back will have a smaller $E_1$ value on its return, thus $E_1$ can only decrease. Similarly it is possible to show that $E_2$ is also always decreasing, for any scenario. All in all, even though the energy can jumps when crossing a switching surface, it is globally decreasing since both $E_1$ and $E_2$ can only decrease. Hence the system is stable and will converge to the origin. Note that this stability analysis is only applicable to inertial loads with linear damping. With a different type of load, this controller could be unstable or exhibiting undesirable behavior such as rapid back-and-forth switching between the operating modes.

\section{Experiment results and discussion}
\label{ev}

\begin{figure}[htpb]
	\centering
		\includegraphics[width=0.40\textwidth]{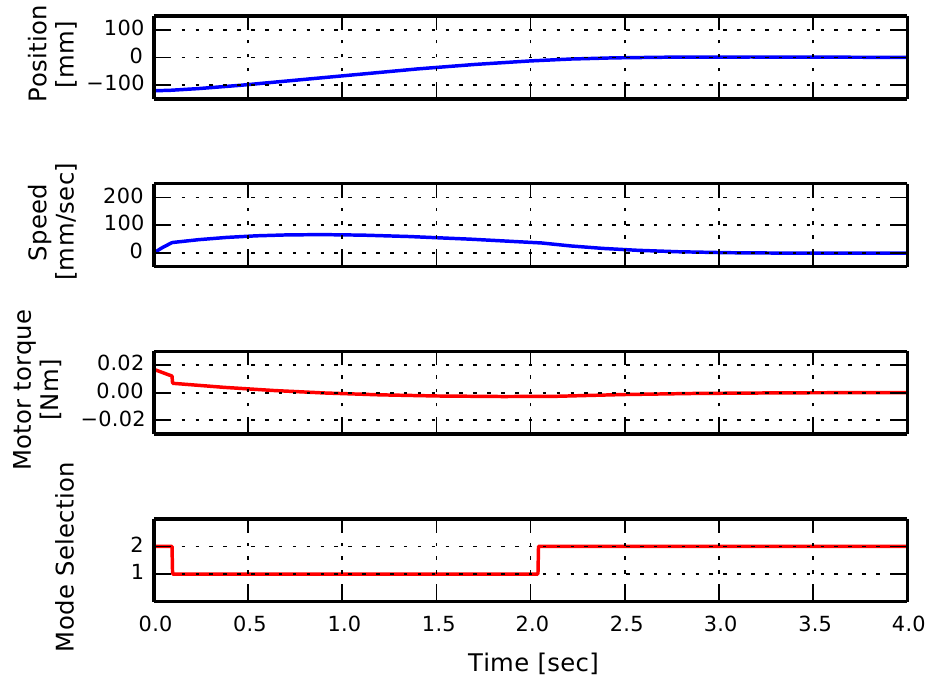}
	\caption{Time trajectory (simulation)}
	\label{fig:time_traj_lqr}
\end{figure}

\begin{figure}[htpb]
	\centering
		\includegraphics[width=0.40\textwidth]{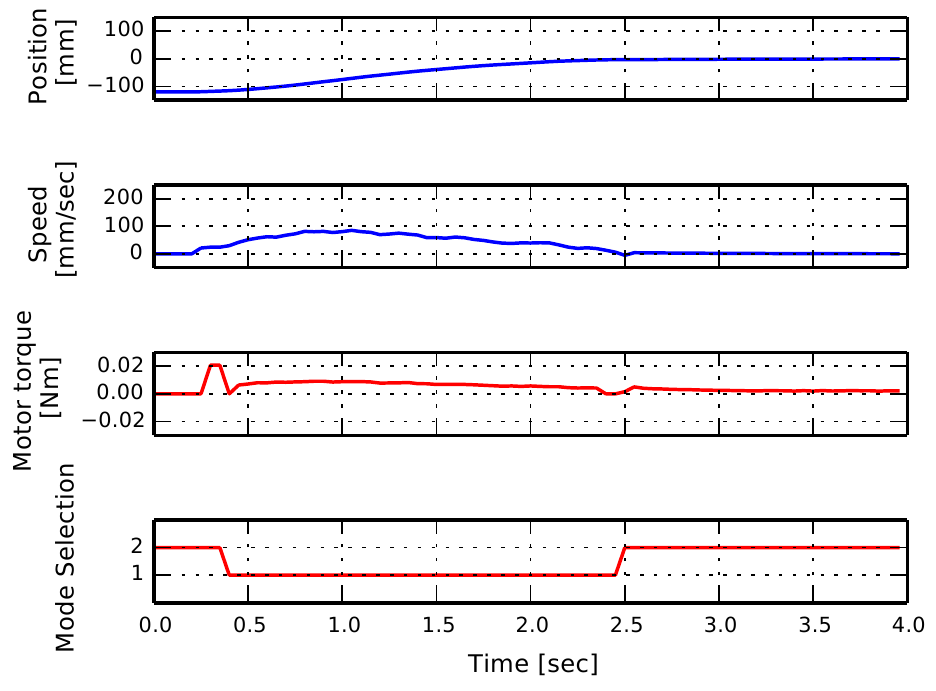}
	\caption{Time trajectory (experiment with prototype)}
	\label{fig:optimal_experiment}
\end{figure}

A trajectory of the system using the approximated control law of eq.\eqref{eq:uhat} starting from rest at $x$ = -120 mm is simulated using the simplified model and results are illustrated on Fig. \ref{fig:phase_plot_lqr} and Fig. \ref{fig:time_traj_lqr}. The same experiment is conducted using the real actuator prototype and results are illustrated on Fig. \ref{fig:optimal_experiment}. On both experiments it is possible to see clearly that the controller actions are more aggressive when using the large reduction ratio, which lead the desired optimal non-linear behavior, as discussed in section \ref{sec:NumericalResults}.

The experiment with the real prototype shows that even though the control policy was generated with a simplified linear model, the desired global behavior is still reached. The discrepancy between the simulation and the experiment is mainly due to the fact that the model used in the simulation neglect the transitions dynamic, and also because the linear damping used in the model does not describes very well the friction in the lead screw. Diverse experiments, shown in the attached video, demonstrate that the proposed controller takes advantage of the mode selection for a wide range of situations. Even though the controller is not the optimal policy anymore in those other scenarios, optimizing for an archetype point-to-point task is found to be a practical way to tune global feedback laws that take advantage of the mode selection.

\section{CONCLUSION AND OUTREACH}

In this paper, a practical approach based on dynamic programming is proposed to generate nearly optimal feedback laws for robotic systems with both continuous and discrete input variables. Optimal control policies for point-to-point motions with a two-speed actuator prototype are derived numerically for minimizing the time to arrival, the wasted energy or a quadratic cost function. The optimal feedback policy for the quadratic cost is then approximated to a simple piece-wise analytical function, global stability is proven and performance is demonstrated with the prototype. The main advantages of the technique are: the possibility to include any kind of nonlinearities and constraints (such as discrete input variables), and the generation of analytical feedback laws. Limitations are that approximations must be used for both the system dynamic and the feedback policy function. Next, to follow this initial work on a single actuator, the control of robots using multiple two-speed actuators will be studied. The main challenge will be to find good approximations so that the problem is still tractable computationally.



\bibliographystyle{src/IEEEtran}
\bibliography{main}

\end{document}